\documentclass[runningheads]{llncs}
\usepackage{amsmath,amssymb,amsfonts}
\expandafter\let\csname equation*\endcsname\relax
\expandafter\let\csname endequation*\endcsname\relax
\usepackage{algorithmic}

\usepackage[T1]{fontenc}
\usepackage{graphicx}

\usepackage{booktabs}
\usepackage{array}
\usepackage{tabularx}
\usepackage{multirow}
\usepackage{multicol}       
\usepackage{threeparttable}
\usepackage{makecell}
\usepackage{enumitem}

\usepackage{cite}
\usepackage{url}
\usepackage{orcidlink}
\usepackage{hyperref}
\usepackage{color}

\definecolor{main}{HTML}{adbde3}   
\definecolor{sub}{HTML}{ebebeb}   

\usepackage{siunitx}
\usepackage{setspace}             

\begin{document}
\title{Fourier Neural Operators for Rayleigh–Bénard Convection}
%
%

\author{Chelsea Maria John\inst{1,2}\orcidID{0000-0003-3777-7393}
\and Thibaut Lunet\inst{2}\orcidID{0000-0003-1745-0780}
\and Sebastian G\"otschel\inst{2}\orcidID{0000-0003-0287-2120}
\and Andreas Herten\inst{1}\orcidID{0000-0002-7150-2505}
\and Stefan Kesselheim\inst{1}\orcidID{0000-0003-0940-5752}
\and Daniel Ruprecht\inst{2}\orcidID{0000-0003-1904-2473}}

\authorrunning{C.M. John et al.}

\institute{
J\"ulich Supercomputing Centre, J\"ulich, Germany\\
\email{\{c.john,a.herten,s.kesselheim\}@fz-juelich.de}
\and
Hamburg University of Technology, Hamburg, Germany\\
\email{\{thibaut.lunet,sebastian.goetschel,ruprecht\}@tuhh.de}
}
\maketitle              
\begin{abstract}
We propose an improved Fourier Neural Operator (FNO) for modeling two-dimensional Rayleigh–Bénard convection by predicting time increments instead of full solutions, achieving higher accuracy than a standard FNO baseline. The resulting model is compact (314k parameters, \qty{1.26}{\mega\byte}) and fast (\qty{7}{\milli\second} inference), while maintaining similar accuracy as demonstrated in previous benchmarks. We show that although FNOs generalize to finer meshes, accuracy remains limited by the resolution of the training data.
\end{abstract}

\section{Introduction}
Modeling turbulent convection is challenging and has applications from atmospheric flows~\cite{atmosphereRBC} to industrial processes such as silicon wafer production~\cite{WaferCFD}. A standard benchmark is Rayleigh–Bénard convection (RBC), where a fluid heated from below develops an overturning circulation~\cite{turbulentHeatFlows} with the strength of turbulence governed by the Rayleigh number ($Ra$). 
At high Ra the flow becomes strongly turbulent, making high-resolution numerical simulations computationally expensive. 
Recent machine learning approaches, particularly Fourier Neural Operators (FNOs)~\cite{li2021fourierneuraloperatorparametric}, provide an alternative to mesh-based solvers by learning solution operators that, unlike PINNs~\cite{cai2021physicsinformedneuralnetworkspinns}, generalize across resolutions. In this work, we apply FNOs to 2D RBC using Dedalus~\cite{dedalus} as ground truth, focusing on stable small-step predictions that preserve turbulent statistics and align with time-stepping solvers.
We propose a lean FNO design suitable for use in iterative methods such as Parareal~\cite{parareal,IbrahimEtAl2025}.

\paragraph{Related Work.}
Hammoud et al.~\cite{PinnBackward} show that PINNs reconstruct previous RBC states with relative $L_2$-errors on the order of $10^{-1}$ at $\text{Ra}=10^7$, while Almeida et al.~\cite{encoderdecoderDeepOnet} achieve less than \qty{10}{\percent} reconstruction error using an encoder–decoder DeepONet-based approach~\cite{DeepOnet}. 
Wu et al.~\cite{wu2025tantetimeadaptiveoperatorlearning} introduce COAST, a causal operator framework with adaptive time stepping, reporting a VRMSE of 0.23 for $Ra=10^7$ over a 5-step rollout. 
Straat et al.~\cite{Straat2025-mn} apply FNO to 2D RBC at multiple Rayleigh numbers and obtain an average relative $L_2$-error of $0.021$ on a $96 \times 64$ grid. 
In contrast, our 2D lean FNO model learns to predict increments rather than solutions and achieves significantly higher accuracy when predicting future solution states.
It has fewer parameters and faster inference, which is beneficial when coupled with iterative numerical algorithms. 
However, it exhibits faster error accumulation under longer auto-regressive rollouts or larger time steps compared to 3D FNOs.

\paragraph{Contributions}
We propose a lean FNO architecture that learns the \emph{increment} from the state at time $t$ to a state at time $t + \Delta t$, similar to a time-stepping algorithm in a mesh-based simulation.
We perform an ablation study which reveals that using multi-layer 1D convolutional scaling operators delivers better accuracy than linear layers.
The performance of our lean FNO architecture in terms of accuracy, memory footprint and inference times is compared against the model by~\cite{Straat2025-mn}.
Finally, we analyze how FNO generalizes across spatial and temporal resolutions and show that it interpolates to new meshes, but that accuracy remains limited by the resolution of the training data.

\section{Rayleigh-Bénard Convection}

Rayleigh-Bénard convection occurs when a fluid within a confined channel is subjected to a temperature difference. 
In our setup, the lower plate is heated to a temperature $\theta_{H}$ while the upper plate is cooled to $\theta_{C}$ with a vertical distance (d) between the plates. %
This causes the cooler, denser fluid to sink and the warmer, lighter fluid to rise, generating convective rolls whose patterns are influenced by the aspect ratio of the channel and the physical properties of the fluid. 
The non-dimensional momentum conservation equations governing the dynamics of RBC are
\begin{equation}
    \label{eq:u_t}
    \frac{\partial \mathbf{u}}{\partial t} + (\mathbf{u} \cdot \nabla) \mathbf{u} = -\nabla p + \theta \mathbf{\hat{z}} + \sqrt{\frac{\text{Pr}}{\text{Ra}}} \nabla^2 \mathbf{u}. 
\end{equation}
Here, $\mathbf{u} = (u, w)$ denotes the velocity vector with horizontal component $u$ and vertical component $w$, $p$ is the pressure, $\theta$ is buoyancy, and the Prandtl number $\text{Pr}$ is the ratio of kinematic viscosity $\mu$ and thermal diffusivity $\kappa$. 
The Rayleigh number $\text{Ra}$, which characterizes the convective motion, is defined as
\begin{equation}
    \text{Ra} = \frac{\alpha \text{g} (\theta_{H}-\theta_{C})d^3}{\mu \kappa}
\end{equation} 
with $\alpha$ being the thermal expansion coefficient and $\text{g}$ the acceleration due to gravity.
The continuity equation for incompressible flow that ensures mass conservation is
\begin{equation}\label{eq:divu}
    \nabla \cdot \mathbf{u} = 0
\end{equation}
Conservation of energy provides the following equation for the temperature
\begin{equation}
    \label{eq:theta_t}
    \frac{\partial \theta}{\partial t} + (\mathbf{u} \cdot \nabla) \theta = \frac{1}{\sqrt{\text{Ra Pr}}}\nabla^2 \theta.
\end{equation}

\section{Fourier Neural Operator}
Fourier Neural Operators (FNOs) solve PDEs by learning mappings between function spaces using spectral representations~\cite{li2021fourierneuraloperatorparametric}. The architecture consists of an input lifting layer $\mathcal{P}$ that maps the input function $a$ from channel dimension $d_a$ to a higher-dimensional representation $v$ of dimension $d_v$, followed by multiple Fourier layers and a final projection $\mathcal{Q}$ that maps to the output $u$ with channel dimension $d_u$. Each Fourier layer applies a Fourier transform $\mathcal{F}$, performs spectral convolution on a fixed set of low-frequency modes $R$, and transforms back via the inverse transform $\mathcal{F}^{-1}$. A local linear term $Wv$ and a nonlinear activation $\sigma$ are added to capture local and higher-frequency features beyond the truncated spectrum.

\subsection{Generation of training data}\label{sec:data}
Training and evaluation data were generated using Dedalus~\cite{dedalus} on a $256 \times 64$ grid with a $RK443$ time stepper and $\Delta t = 10^{-3}$, with $\theta_H=1$ and $\theta_C=0$. 
Simulations start from random perturbations, run to $T_{init}=100$ to reach a pseudo-steady turbulent state.
Data is collected from $T_{init}=100$ to $T_{sim}=200$ for $\text{Pr}=1$ and $\text{Ra}=10^7$, with spectral analysis confirming adequate resolution. 
Ten simulations with different random seeds are used to record $\mathcal{U}(t)=[\mathbf{u}(t),\theta(t),p(t)]$ at each time step, forming 200 input–output pairs $(\mathcal{U}(t),\mathcal{U}(t+\Delta t))$ over $[100,200]$. 
From 2000 samples, \qty{80}{\percent} are used for training and \qty{20}{\percent} for validation.

\subsection{Learning Objective}\label{sec:obj}
FNOs are typically trained to predict the full solution of a PDE, but learning updates to the state can be more effective. We therefore compare two objectives: a \textbf{solution objective}, where the model directly predicts $\mathcal{U}(t+\Delta t)$ from $\mathcal{U}(t)$, and an \textbf{increment objective}, where it predicts the scaled update $\Delta \mathcal{U} = \Delta t^{-1}(\mathcal{U}(t+\Delta t)-\mathcal{U}(t))$. In the latter case, the solution is recovered via $\mathcal{U}(t+\Delta t)=\mathcal{U}(t)+\Delta t\,\mathcal{O}(\mathcal{U}(t))$, where $\mathcal{O}(\mathcal{U}(t))$ is the model increment prediction, making the FNO analogous to a one-step time integrator.

Both objectives are trained using a relative $L_2$ loss, applied either to the solution or the increment. For evaluation, we measure the relative error in the reconstructed solution.
Note that loss and error coincide only for the solution objective. Since small $\Delta t$ leads to nearly identical states, trivial predictions can yield low loss; therefore, we compare against an identity baseline that propagates $\mathcal{U}(t)$ unchanged (or predicts zero increment) to ensure meaningful learning beyond this naive predictor.

\section{Results}\label{sec:results}
Training was conducted on a single NVIDIA A100 (40 GB) GPU of the JURECA-DC supercomputer, using a custom codebase~\cite{cfno} and data\footnote{\href{https://huggingface.co/datasets/chelseajohn/FNO-RBC2D_paper_artefacts}{https://huggingface.co/datasets/chelseajohn/FNO-RBC2D\_paper\_artefacts}}. Our ablation study on $\mathcal{P}$ and $\mathcal{Q}$ (linear vs. 1D convolutional layers with $d_v=16$) shows that increasing both depth and width improves performance and using cosine LR scheduler leads to better learning. The improved model is further evaluated to test the temporal and spatial generalization, and extended to compare against existing work.
\subsection{Training}
The baseline model is trained using both the solution and increment objectives for $\Delta t = 10^{-3}$. \autoref{Tab:Obj} compares the relative error for both objectives across the four components of the solution, that is velocity in the horizontal ($u$) and vertical ($w$) directions, buoyancy($\theta$), and pressure ($p$), averaged over 200 samples. Furthermore, the training loss is lower than the identity loss for increment objective but not for the solution objective.

The average relative error for the increment objective is two orders of magnitude smaller than for the solution objective. The noise introduced by the FNO when predicting the solution leads to an increase in error by about a factor of ten compared to when the FNO predicts the increment for buoyancy contours.
Predicting increments substantially improves the quality of the generated solutions.
\begin{table}[ht]
\sisetup{detect-weight,
         output-exponent-marker = e,
}
    \centering
    \caption{Relative error computed for increment and solution objective.}
    \begin{tabular}{lS[table-format = 2.1e1]S[table-format = 2.1e1]S[table-format = 2.1e1]}
    \toprule
    \bf {Variable} & \bf {Increment} & \bf {Solution} & \bf {IdError} \\
    \midrule
    $u$ &       5.7e-05 & 1.6e-03 & 2.0e-04\\ 
    $w$ &       9.3e-05 & 1.7e-03 & 1.9e-04\\ 
    $\theta$ &  6.0e-05 & 1.0e-03 & 1.1e-04\\ 
    $p$ &       2.4e-05 & 1.4e-03 & 7.2e-05\\
    \midrule
    {Average} &\bfseries 5.8e-05 & 1.4e-03 & 1.4e-04 \\ 
    \bottomrule
    \end{tabular}
 \label{Tab:Obj}
\end{table}
\begin{table}[ht]
\centering
\caption{Comparison of the improved and baseline FNO configuration}
\begin{tabular}{p{0.55\textwidth}p{0.15\textwidth}p{0.15\textwidth}}
    \toprule
    \textbf{Component} & \textbf{Improved} & \textbf{Baseline} \\
     & \textbf{Model} & \textbf{Model} \\
    \midrule
    \textbf{Objective} & Increment & Solution\\
    \textbf{Kernel} & FNO & FNO \\
    \textbf{Activation Function} ($\sigma$) & GELU & GELU \\
    \textbf{Optimizer} & Adam & Adam \\
    \textbf{LR Scheduler} & Cosine & StepLR \\
    \textbf{Fourier Layers} & 2 & 2 \\
    \textbf{Fourier Modes} & 12 & 12 \\
    \textbf{Scaling Layer ($\mathcal{P}$,$\mathcal{Q}$)} & 4$\times$1D Conv \small (width=$4d_v$) & 1$\times$Linear \small (width=$d_v$) \\
    \textbf{Input Channels} ($d_a$) & 4 & 4 \\
    \textbf{Projection Channels} ($d_v$) & 16 & 16 \\
    \textbf{Output Channels} ($d_u$) & 4 & 4 \\
    \textbf{Total Parameters} & 314772 & 295552 \\
    \textbf{Model Size (MB)} & 1.26 & 1.18 \\
    \textbf{Inference time for batchsize=1 (ms)} & 7 & 5 \\
    \textbf{Relative Error in} $u$ & 2.4e-05 & 1.6e-03 \\
    \textbf{Relative Error in} $w$ & 3.2e-05 & 1.7e-03 \\
    \textbf{Relative Error in} $\theta$ & 1.8e-05 & 1.0e-03 \\
    \textbf{Relative Error in} $p$ & 8.5e-06 & 1.4e-03 \\  
    \textbf{Average error} & \bf 2.1e-05 &  1.4e-03 \\
    \bottomrule
\end{tabular}
\label{Tab:ConfigFNO}
\end{table}

\subsection{Improved model}\label{subsec:improved_model}
Based on the results above, we propose the model configuration described in \autoref{Tab:ConfigFNO} as improvement over the baseline FNO. 
We train both for \num{11500} epochs to predict increments $\Delta t = 10^{-3}$ using data generated as mentioned in \autoref{sec:data} which takes 11h on a single NVIDIA A100 GPU.
\autoref{Tab:ConfigFNO} also shows the relative error in the four variables in~\eqref{eq:u_t} and~\eqref{eq:theta_t} against the Dedalus reference on $ 256 \times 64$ grid for 200 samples from the validation dataset when predicting a single increment $\Delta t = 10^{-3}$.
\subsection{Impact of model time step and autoregressive FNO application}
To evaluate the impact of different time steps, the model is retrained for $\Delta t = 1$, $10^{-1}$, $10^{-2}$, in addition to $10^{-3}$ using the same architecture (\autoref{Tab:ConfigFNO}) on \num{2000} samples for \num{11500} epochs. Models are evaluated over time horizons of $1$, $0.1$, $0.01$, and $0.001$ starting from $t=100$, using autoregressive rollout when the model time step is smaller than the target horizon. As shown in \autoref{Tab:timesteps}, shorter horizons yield lower validation errors due to reduced solution change, while autoregressive inference introduces mild error accumulation, leading to slightly higher errors than single-step predictions at matching time steps. 

\begin{table}[ht]
    \setlength{\tabcolsep}{4.5pt}
    \sisetup{detect-weight,
             output-exponent-marker = e,
    }
    \centering
    \caption{Relative error under temporal generalization.}
    \begin{tabular}{SSSS[table-format = 2.1e1]S[table-format = 2.1e1]S[table-format = 2.1e1]S[table-format = 2.1e1]S[table-format = 2.1e1]}
    \toprule
    \textbf{Data} & \textbf{Model}    & \textbf{Model} & {$u$} & {$w$} & {$\theta$} & {${p}$} & \textbf{Average} \\
    \textbf{Timestep} & \textbf{Timestep} & \textbf{Steps} & &  & & & \textbf{Error}\\
    \midrule
     1     &  0.001 &  1000 & 8.7e-02  & 9.8e-02 & 5.0e-02 & 4.0e-02 & 6.9e-02 \\
     0.1   &  0.001 &  100  & 8.5e-03  & 1.2e-02 & 7.4e-03 & 3.7e-03 & 8.0e-03 \\
     0.01  &  0.001 &  10   & 8.6e-04  & 1.3e-03 & 7.7e-04 & 3.6e-04 & 8.1e-04 \\
     0.001 &  0.001 &  1    & 8.6e-05  & 1.3e-05 & 7.8e-05 & 3.7e-05 & 8.2e-05 \\
    \bottomrule
    \end{tabular}
 \label{Tab:timesteps}
\end{table}
\begin{table}[th]
\sisetup{detect-weight,
         output-exponent-marker = e,
}
    \centering
    \caption{Relative error against the $(256, 64)$ Dedalus reference when evaluating the FNO trained on $(256, 64)$ across different meshes; the $(512, 128)$ case is compared to its own Dedalus reference.
    The interpolation column shows the error when $(256, 64)$ FNO outputs are upsampled to $(512, 128)$ via FFT.}
    \begin{tabular}{lS[table-format = 2.1e1]S[table-format = 2.1e1]S[table-format = 2.1e1]S[table-format = 2.1e1]S[table-format = 2.1e1]S[table-format = 2.1e1]}
    \toprule
    \textbf{Variable} & 
    \textbf{(64, 64)} & 
    \textbf{(256, 32)}&
    \textbf{(64, 32)} &
    \textbf{(256, 64)} &
    \textbf{(512, 128)} &
    \textbf{Interpolation} \\
    \midrule
    $u$           & 7.9e-04 & 4.9e-04 & 5.3e-04 & 2.4e-05 & 1.0e-04 & 1.1e-4 \\ 
    $w$           & 4.7e-04 & 6.2e-04 & 4.6e-04 & 3.2e-05 & 1.4e-04 & 2.3e-4 \\ 
    $\theta$      & 2.4e-04 & 2.6e-04 & 2.7e-04 & 1.8e-05 & 9.2e-05 & 2.0e-4 \\ 
    $p$           & 2.4e-04 & 3.9e-04 & 6.1e-04 & 8.5e-06 & 4.0e-05 & 4.0e-5 \\
    \midrule
    Average       & 4.4e-04 & 4.4e-04 & 4.7e-04 & \bf 2.1e-05 & 9.4e-05  & 1.5e-4 \\ 
    \bottomrule
    \end{tabular}
    \label{Tab:grids}
\end{table}
\subsection{Resolution invariance}
To evaluate generalization across mesh resolutions, the FNO trained on $\Delta t = 10^{-3}$ increments is tested on a $512 \times 128$ grid, twice the training resolution of $256 \times 64$ used for $Ra = 10^7$. 
Errors are computed against reference Dedalus solutions and are found to be of the same order as those on the $256 \times 64$ grid~(\autoref{Tab:grids}). 
Errors change very little if the reference solution on a $512 \times 128$ is restricted to a $ 256 \times 64$ mesh and this data used as input to the FNO. 
Evaluating the FNO on the trained resolution ($ 256 \times 64$) and interpolating the output to the finer grid via FFT delivers very similar errors as evaluating the FNO directly on the finer mesh.
This demonstrates that the FNO shows mesh invariance by interpolating to unseen resolutions without significant additional error. 
However, unlike numerical solvers, increasing inference resolution does not improve accuracy, which is instead determined by the resolution of the training data.

\subsection{Comparison with Straat et al.}
To evaluate the performance of our improved FNO architecture, we compare it against the model of Straat et al.~\cite{Straat2025-mn} for $Ra = 5 \times 10^6$ with $Pr = 0.7$ on a $96 \times 64$ grid and boundary temperatures $\theta_H = 2$ and $\theta_C = 1$. We adopt their setup but use an architecture with 8 Fourier layers, 64 projection channels, and 16 Fourier modes to better capture the turbulent dynamics and larger time step. The model is trained on \num{44940} samples from 15 Dedalus simulations and validated on \num{14980} samples from 5 simulations, each run for \qty{300}{\second} with data sampled at $\Delta t = 0.1$ after a \qty{200}{\second} warm-up, and trained on JURECA-DC using 8 NVIDIA A100 GPUs for \num{500} epochs in 3.5 hours, reaching a training and validation loss of \num{0.03}.

For evaluation, predictions are generated over a \qty{30}{\second} window via 60 recursive steps with a model time step of \qty{0.5}{\second}. Averaged over 50 samples from 10 random starting points across 5 validation runs, our model achieves an average loss of $0.11$ (Table~\ref{Tab:straat_model}), compared to $0.04$ reported by Straat et al. Direct component comparison is not possible due to lack of data in Straat et al. While their FNO3D accumulates error more slowly, our model attains comparable initial accuracy despite its much smaller architecture, requiring only 132 MB (33 million, FP32 parameters) versus 3037 MB, and achieves faster inference times of 10 ms (batch size 10) and 30 ms (batch size 50) per prediction window, compared to 0.45 s on an NVIDIA A40 for Straat et al.
\begin{table}[ht]
\sisetup{
  detect-weight,
  output-exponent-marker = e,
}
\centering
\caption{Relative error solving the Straat et al. benchmark.}
\begin{tabular}{
    l
    S[table-format=1.1e-1]
    S[table-format=1.1e-1]
    S[table-format=1.1e-1]
    S[table-format=1.1e-1]
}
\toprule
\bf Variable & \multicolumn{2}{c}{\bf $t=0.5$}& \multicolumn{2}{c}{\bf $t=30\,\mathrm{s}$} \\
\cmidrule(lr){2-3}\cmidrule(lr){4-5}
  & \textbf{Error} & \textbf{IdError} &\textbf{Error} & \textbf{IdError} \\
\midrule
$u$       & 4.6e-02 & 1.8e-01 & 2.0e-01 & 2.1e-01 \\
$w$       & 4.9e-02 & 1.4e-01 & 1.8e-01 & 2.0e-01 \\
$\theta$  & 2.1e-02 & 3.4e-02 & 3.1e-02 & 3.6e-02 \\
$p$       & 1.3e-02 & 3.0e-02 & 3.5e-02 & 3.7e-02 \\
\midrule
Average   &  \bf 3.2e-02 &  9.8e-02 &  \bf 1.1e-01 &  1.2e-01 \\
\bottomrule
\end{tabular}
\label{Tab:straat_model}
\end{table}

\section{Conclusion}

In this work, we systematically evaluate Fourier Neural Operators for two-dimensional Rayleigh–Bénard convection in the turbulent regime. We find that predicting increments rather than full states significantly improves accuracy and stability, reducing errors by up to two orders of magnitude for small time steps and mitigating noise, effectively turning the FNO into a data-driven one-step integrator. Architectural choices such as deeper 1D convolutional lifting/projection layers and a cosine learning rate scheduler further improve accuracy, yielding a compact and efficient model.

Our results show that FNOs are resolution-invariant in the sense that they interpolate to unseen meshes, but their accuracy is ultimately limited by the training data resolution rather than improved by finer inference grids. Compared to the much larger 3D FNO by Straat et al, our lean model achieves a trade-off between accuracy, memory, and inference speed, with substantially smaller footprint and faster runtime but stronger long-horizon error accumulation. These properties highlight the potential of FNOs as efficient surrogate models and as coarse propagators in space–time parallel methods such as Parareal.

\begin{credits}
\subsubsection{\ackname}
This project received funding from the European High-Performance Computing Joint Undertaking (JU) under grant No.~101118139 (Horizon Europe), with compute time on the GCS Supercomputer JURECA at JSC provided by the Gauss Centre for Supercomputing e.V.

\end{credits}

%
%
%
\bibliographystyle{splncs04}
\bibliography{references}

\vspace{12pt}

\end{document}